\documentclass{article}

\usepackage{microtype}
\usepackage{graphicx}
\usepackage{booktabs} 

\usepackage{caption}  
\usepackage{subcaption} 
\usepackage[hyphens]{url} 

\usepackage{xcolor} 
\usepackage{soul}  

\usepackage{hyperref}

\usepackage[accepted]{icml2024}

\usepackage{amsmath}
\usepackage{amssymb}
\usepackage{mathtools}
\usepackage{amsthm}

\usepackage{soul} 

\usepackage[capitalize,noabbrev]{cleveref}

\theoremstyle{plain}

\theoremstyle{definition}

\theoremstyle{remark}

\usepackage[textsize=tiny]{todonotes}

\icmltitlerunning{Position: Do Not Explain Vision Models Without Context}

\begin{document}

\twocolumn[
\icmltitle{Position: Do Not Explain Vision Models Without Context}




\begin{icmlauthorlist}
\icmlauthor{Paulina Tomaszewska}{yyy}
\icmlauthor{Przemys{\l}aw Biecek}{yyy,xxx}
\end{icmlauthorlist}

\icmlaffiliation{yyy}{Faculty of Mathematics and Information Science, Warsaw University of Technology, Warsaw, Poland}
\icmlaffiliation{xxx}{Faculty of Mathematics, Informatics and Mechanics, University of Warsaw, Warsaw, Poland}

\icmlcorrespondingauthor{Paulina Tomaszewska}{paulina.tomaszewska111@gmail.com}

\icmlkeywords{Machine Learning, ICML}

\vskip 0.3in
]



\printAffiliationsAndNotice{}  

\begin{abstract}
Does the stethoscope in the picture make the adjacent person a doctor or a patient? This, of course, depends on the contextual relationship of the two objects. If it’s obvious, why don’t explanation methods for vision models use contextual information? In this paper, we (1) review the most popular methods of explaining computer vision models by pointing out that they do not take into account context information, (2) show examples of failures of popular XAI methods, (3) provide examples of real-world use cases where spatial context plays a significant role, (4) propose new research directions that may lead to better use of context information in explaining computer vision models, (5) argue that a change in approach to explanations is needed from \textit{where} to \textit{how}.
\end{abstract}

\section{Introduction}
\label{introduction}  
The number of solutions that incorporate Deep Learning (DL) models is rapidly increasing. Recently, a significant change of the mindset in the community is observed. Many researchers agree that we should not only aim for the models with the highest performance but also the ones that behave in a responsible manner. This aspect is approached using eXplainable AI (XAI) techniques. In the work, we focus only on those for an investigation of Computer Vision models. The most attention is attributed to 
visual explanations. There are already many well-established post-hoc methods that generate colorful heatmaps highlighting the key regions in the input image for the model decision-making process. The major types of explanations are: permutation-based (i.e. LIME~\cite{lime}), gradient-based (i.e. Grad-CAM~\cite{grad-cam}, Integrated gradients~\cite{ig}) and propagation-based (i.e. Layer-wise Relevance Propagation~\cite{lrp}, Concept Relevance Propagation~\cite{crp}). 
For a systematic survey on visual XAI methods, please consult a survey~\cite{xai_survey, Holzinger_survey}.

The biggest attention of the XAI community is focused on the aforementioned visual explanations. In 2019, it was assessed that the number of papers on visual explanations was about 4 times bigger than the number of papers on textual and example-based explanations of models used in medical applications. The same trend was anticipated in 2020~\cite{survey_plot_number}. We suspect that this has not changed much.
The existing visual methods provide valuable insights, however, may not be enough. We identify a set of use cases in which they would fail to accurately explain the model's predictions. These are the cases where the ground truth labels depend on the spatial relationships between objects constituting images. These may be observed in critical domains, such as street surveillance systems, autonomous cars, and healthcare. \textbf{This position paper argues that we should take into account contextual information when explaining vision models -- we need spatial XAI.}

\begin{figure*}[hbt]
\centering
\includegraphics[width=0.95\linewidth]{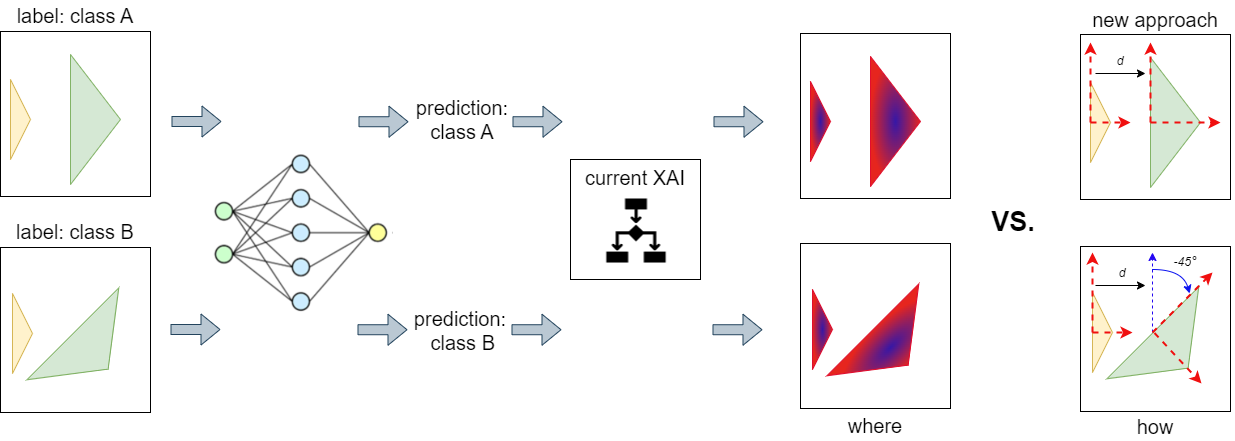}
\caption{Example of two images consisting of the same objects but located differently within the scene (\textit{angular orientation}). A DL model would correctly classify the images into separate classes. However, the common heatmap-based explanations in the two cases will highlight solely the two same triangles and would not capture the spatial relationships that are crucial factors for correct classification. Even after translating the highlighted regions within heatmaps to semantic concepts as suggested by~\citet{crp} (from \textit{where} to \textit{what)}, we would simply learn that in both cases there are two triangles. Therefore, there will be no difference in explanations that should be able to point out that spatial relationships are the main model's decision factor. That is why we postulate a shift of the paradigm from \textit{where} to \textit{how} so that the spatial relationships of how the objects are oriented towards each other will be captured within XAI methods. }
\label{fig:graphical_abstract}
\end{figure*}

\paragraph{From \textit{where} to \textit{how}} The question is how to investigate whether the model takes into account spatial context. This area of model investigation seems largely unexplored.
Recently, \citet{crp} proposed to change the~paradigm in XAI from \textit{where} to \textit{what}. It means that instead of simply highlighting the regions in the input images that are key for the model's prediction, we should focus on extracting \textit{what} semantic features within the images are important. We propose to shift the approach from \textit{where} to \textit{how} so that instead of only operating on the image pixel space \textit{where} the pixels are highlighted, we should also analyze \textit{how} the objects within images are oriented towards each other in the space (Figure~\ref{fig:graphical_abstract}). 
The spatial relationships may be thought of as a kind of analogy to interaction terms that are investigated in models for tabular data analysis.

As the field of XAI matures, we should focus on more complex concepts to capture the ambiguity of data and the~intricate reasoning of DL models. Humans when analyzing images often take into account the composition of the scene. This information can be captured using eye trackers. The~resulting saliency maps are widely studied and modeled using neural networks~\cite{saliency_map_human}.
Therefore, investigating \textit{how} may not only help in getting a deeper understanding of DL models, increasing safety by preventing inaccurate model predictions as we will have XAI tools to evaluate models before the launch but also in checking if there is an alignment between the way human and DL models reason about the scenes.

The structure of the work is as follows. First, we define the~types of contextual information within images and provide illustrative examples (Section~\ref{sec:context}). Next, we elaborate on the domains when spatial context matters (Section~\ref{sec:application}). Then, we provide a summary of works on DL models where the~spatial context plays a central role (Section~\ref{sec:model_context}). We provide examples of failures of popular XAI techniques in Section~\ref{sec:failures}. Later, we contrast the number solutions addressing the topic of spatial context developed in the modelling community with the number of papers in the topic of spatial XAI (Section~\ref{sec:xai}). Lastly, we outline the necessary research directions (Section~\ref{sec:directions}) and draw conclusions (Section~\ref{sec:conclusions}). 

\section{Contextual Information within Images}
\label{sec:context}
The importance of context within input data is vastly studied in Time Series and Natural Language Processing (NLP), yet much less explored in Computer Vision. 

The key elements of scene compositions when analyzing images using DL models were defined by ~\citet{context_survey}: \textit{semantic context}, \textit{spatial context} and \textit{scale
context}. This nomenclature was built upon the taxonomy proposed in a psychological study~\cite{scene_perception}.

\subsection{Semantic Context}
It refers to the situation when the co-presence of the objects in an image is helpful when performing a task.
\subsubsection{Likelihood}
 Here, the fact that objects are within the same space (regardless of spatial relations between them) is an important clue. Suppose there is an image of an empty room with one chair and the task is to classify the room. It seems that it can be a doctor's room, a classroom or a stage in a theatre etc. Without any contextual information, that would be just a pure guess. However, if there is a blackboard in the image, it becomes clear that most probably it is a classroom. In this case, the information on the co-presence of other objects in the scene makes a classification task easier.

\begin{figure}[hbt]
\centering
\includegraphics[width=0.95\linewidth]{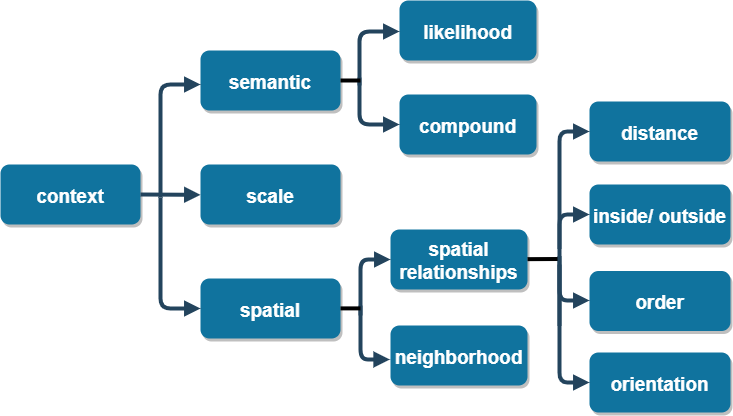}
\caption{Taxomony of contextual information within images. It~is~an~extended version of the one proposed by~\citet{context_survey}.}
\label{fig:taxonomy}
\end{figure}

\subsubsection{Compound}
We would like to add another category of semantic context that was not pointed out by~\citet{context_survey}: \textit{compound}. 
There are objects i.e. rollerblades that are a~compound of other self-existing objects (here, shoes and wheels). The information on whether there are wheels attached to shoes is crucial to classify the object correctly -- shoes without wheels are not rollerblades anymore. Let us consider the possible outcome of the existing XAI methods when explaining the rollerblades prediction. When a model is only supposed to classify shoes vs rollerblades then it may be sufficient for the model to focus only on the detection of wheels and therefore this is the part that most probably will be highlighted within the explanation heatmap. However, if there is a classification of rollerblades vs. cats then we would rather expect to see in the heatmap both the shoes and wheels highlighted. Therefore, we may face some relativity of explanations.

\begin{figure*}[hbt!]
\centering
\begin{subfigure}[b]{0.22\linewidth} %
\centering
\includegraphics[width=0.8\linewidth]{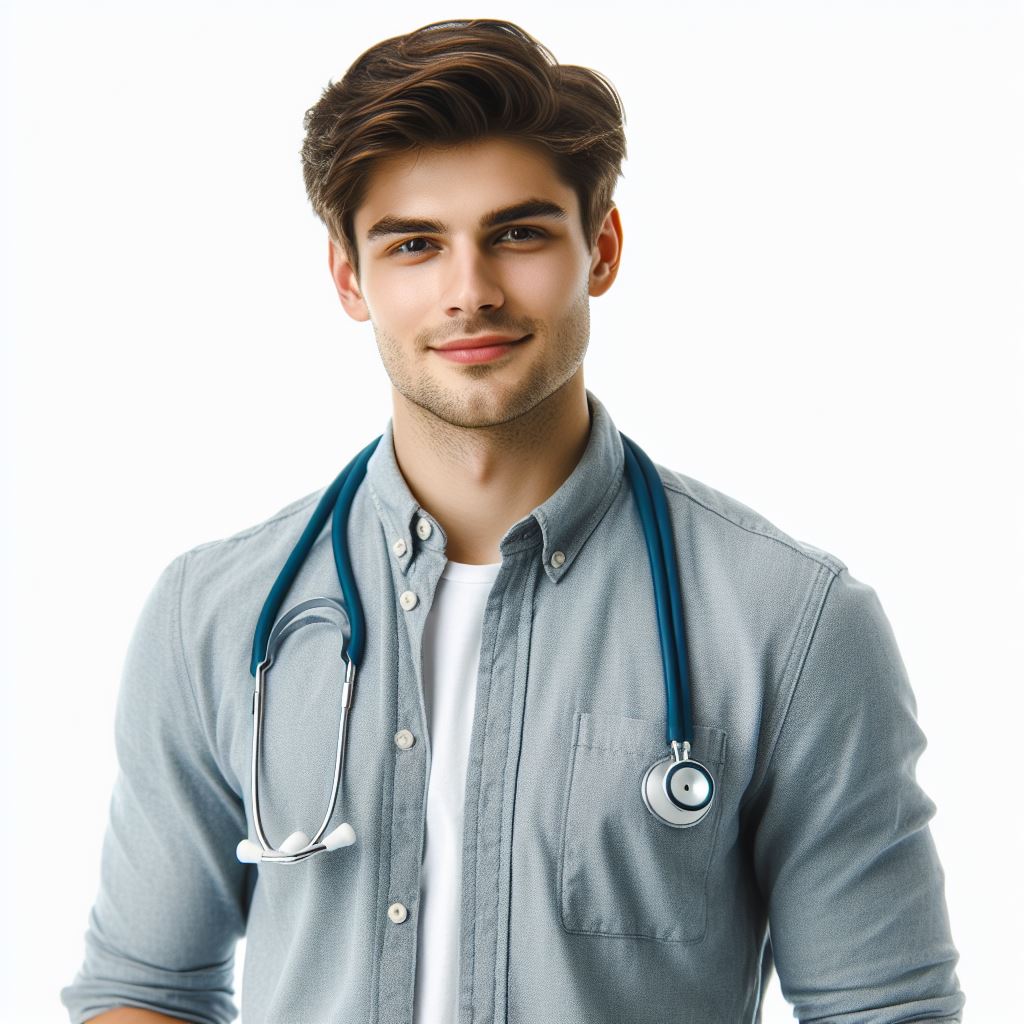}
    \caption{physician}
    \label{distance1}
    \end{subfigure} 
\begin{subfigure}[b]{0.22\linewidth} %
\centering
\includegraphics[width=0.8\linewidth]{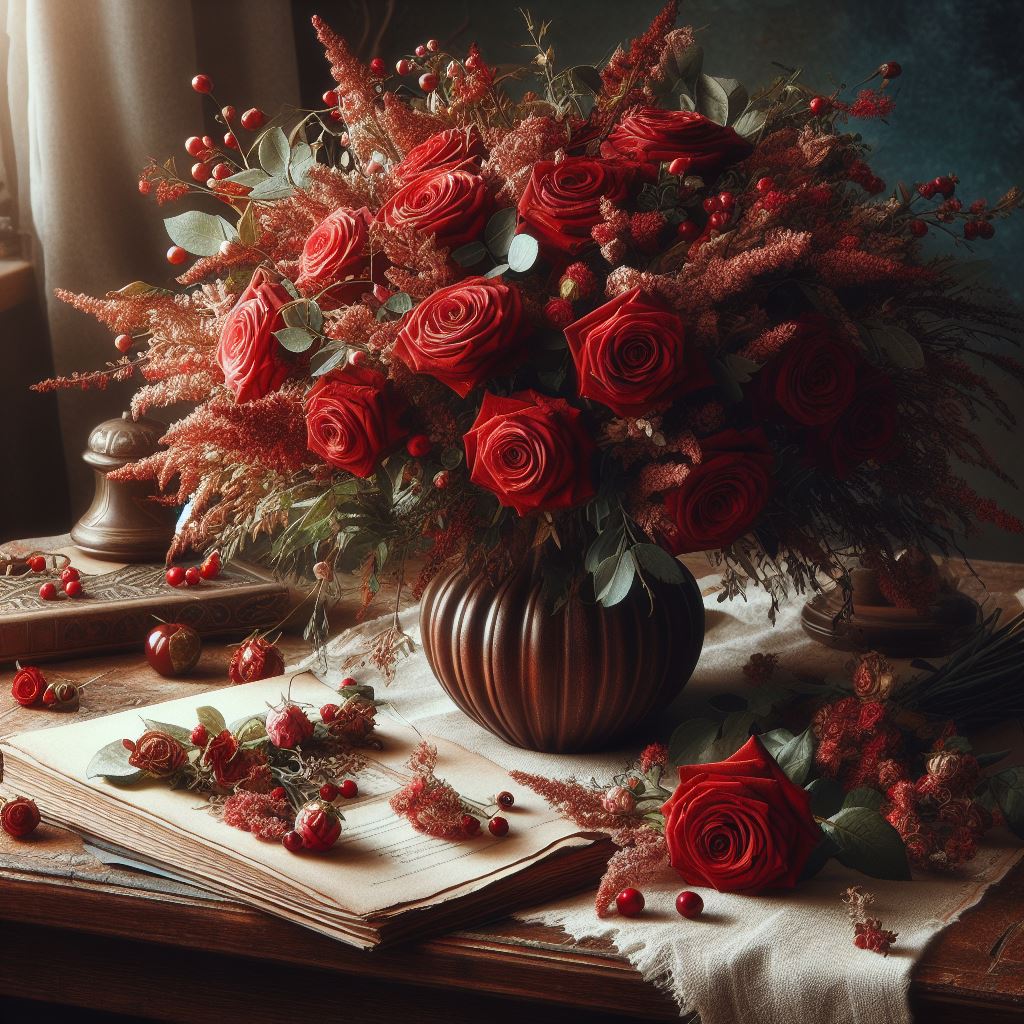}
\caption{flowers in a vase}
\label{inside1}
\end{subfigure} 
\begin{subfigure}[b]{0.22\linewidth} %
\centering
\includegraphics[width=0.8\linewidth]{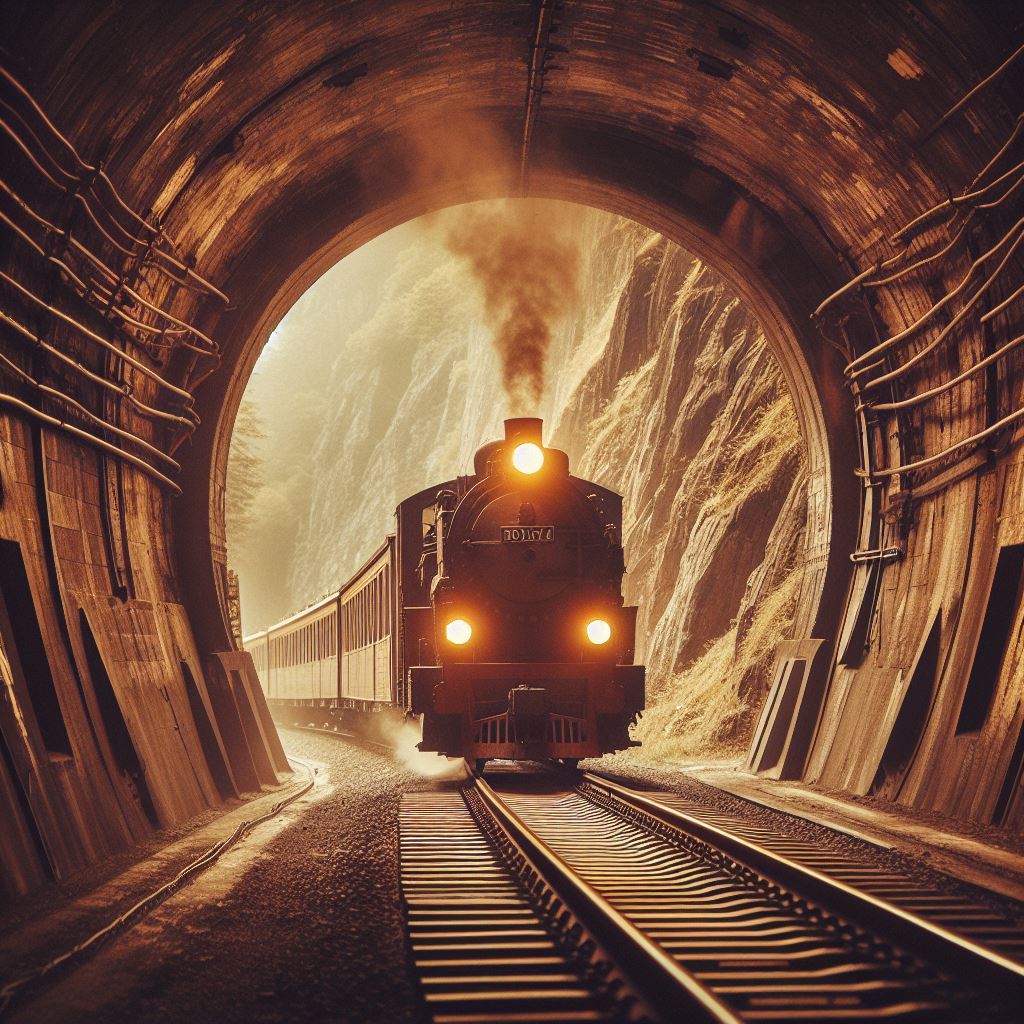}
\caption{train entering a tunnel}
\label{order1}
\end{subfigure} 
\begin{subfigure}[b]{0.22\linewidth} %
\centering
\includegraphics[width=0.8\linewidth]{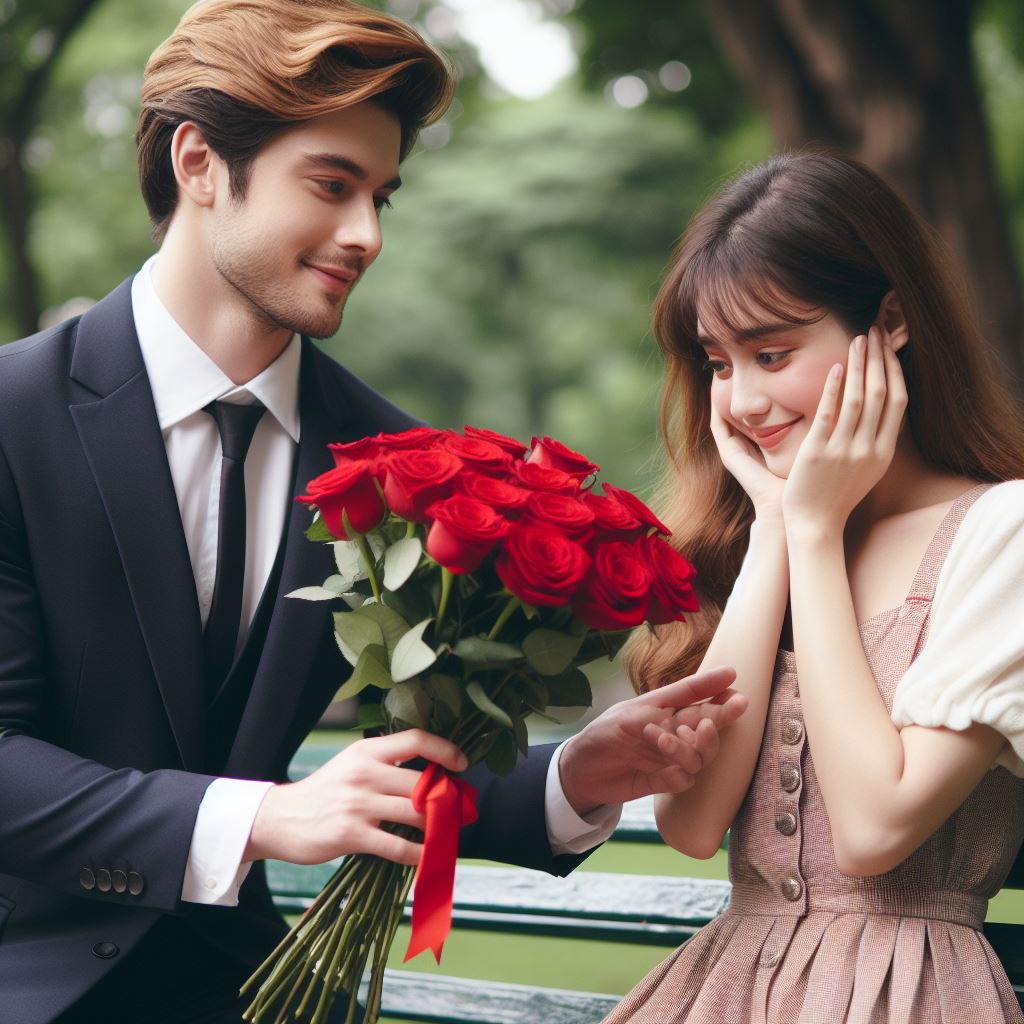}
\caption{man giving roses}
\label{orientation1}
\end{subfigure}   
\hfill
\begin{subfigure}[b]{0.22\linewidth} %
\centering
\includegraphics[width=0.8\linewidth]{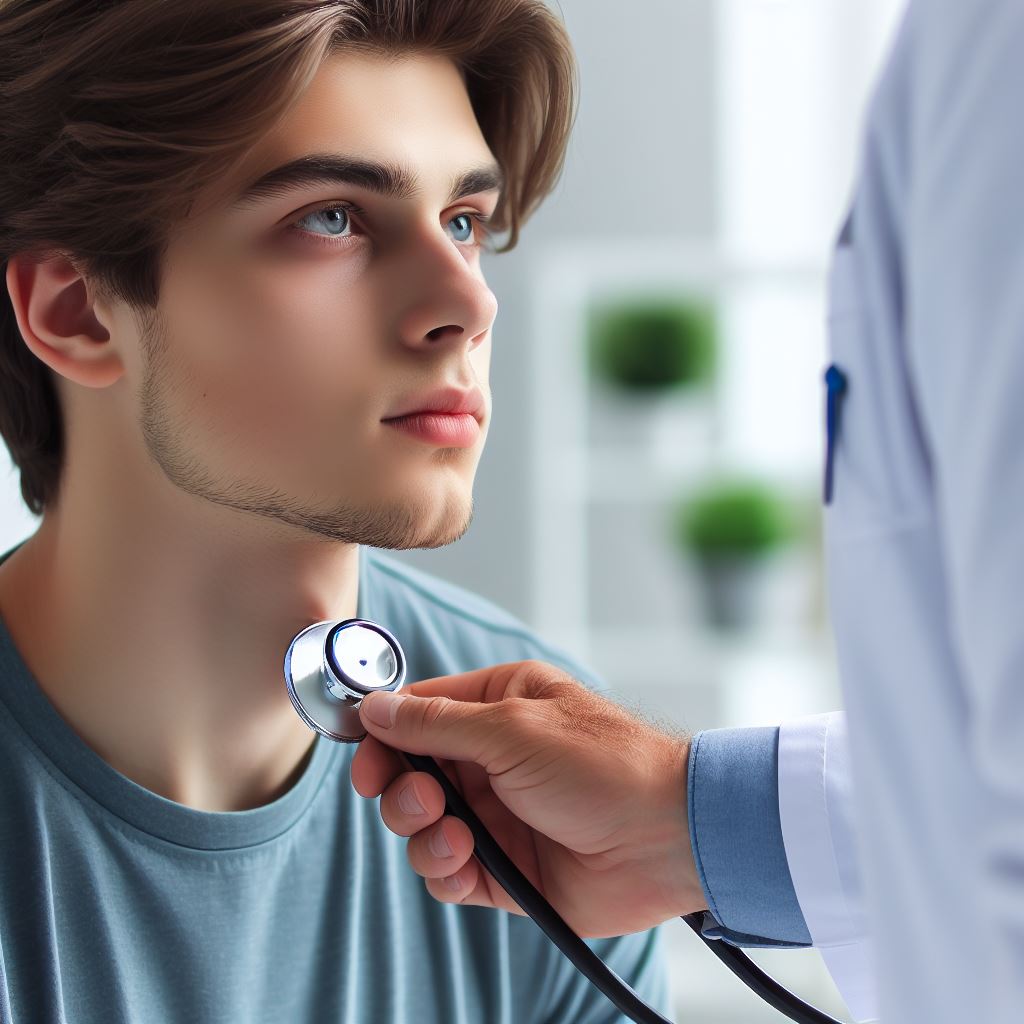} 
\caption{patient}
\label{distance2}
\end{subfigure}
\begin{subfigure}[b]{0.22\linewidth} %
\centering
\includegraphics[width=0.8\linewidth]{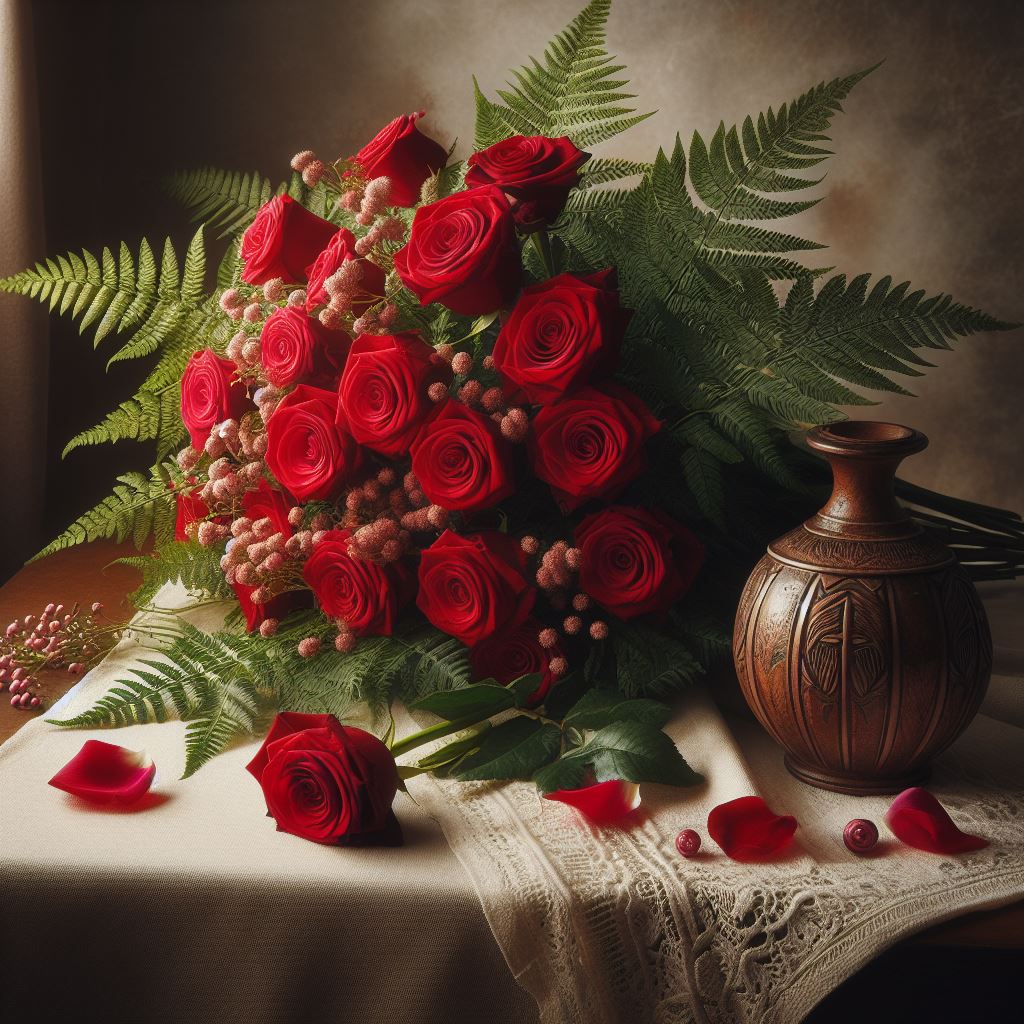}
\caption{flowers near vase}
\label{inside2}
\end{subfigure} 
 \begin{subfigure}[b]{0.22\linewidth} %
\centering
\includegraphics[width=0.8\linewidth]{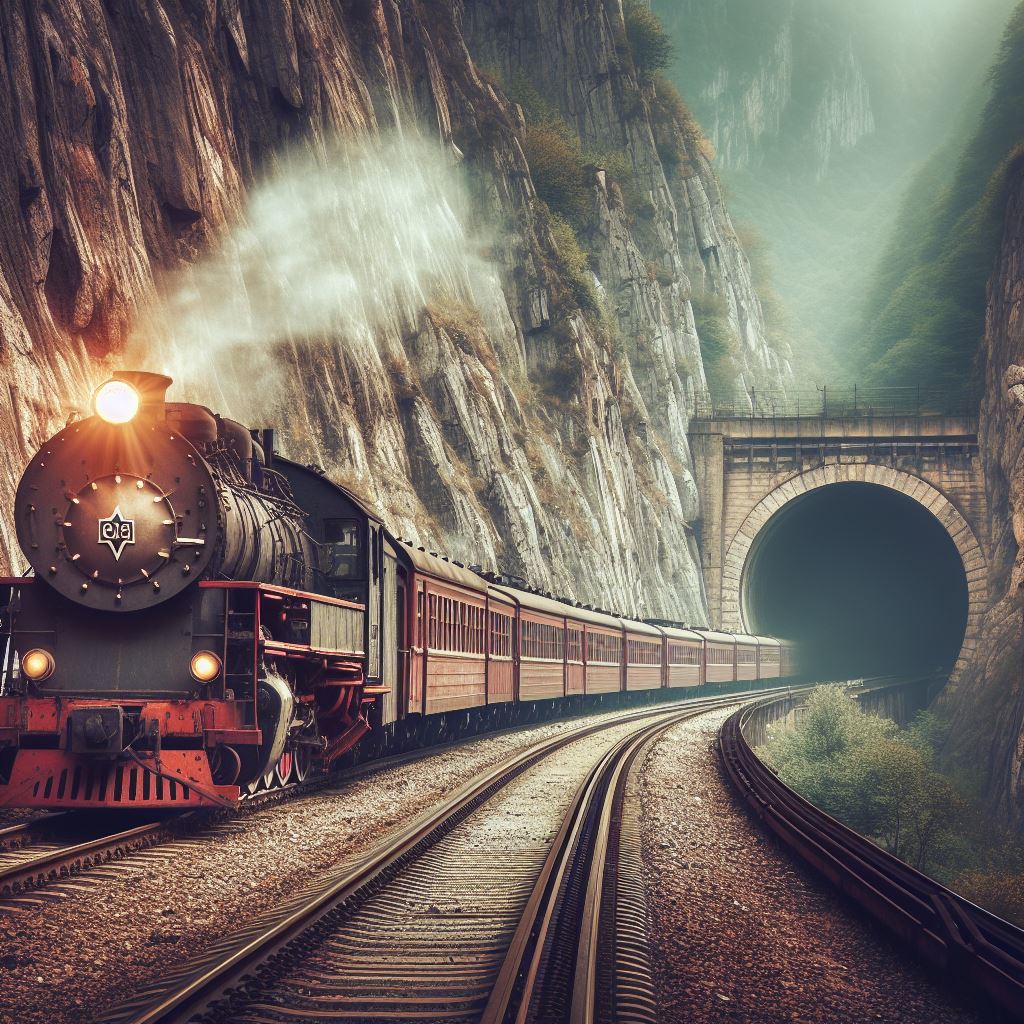}
\caption{train leaving a tunnel}
\label{order2}
\end{subfigure}    
\begin{subfigure}[b]{0.22\linewidth} %
\centering
\includegraphics[width=0.8\linewidth]{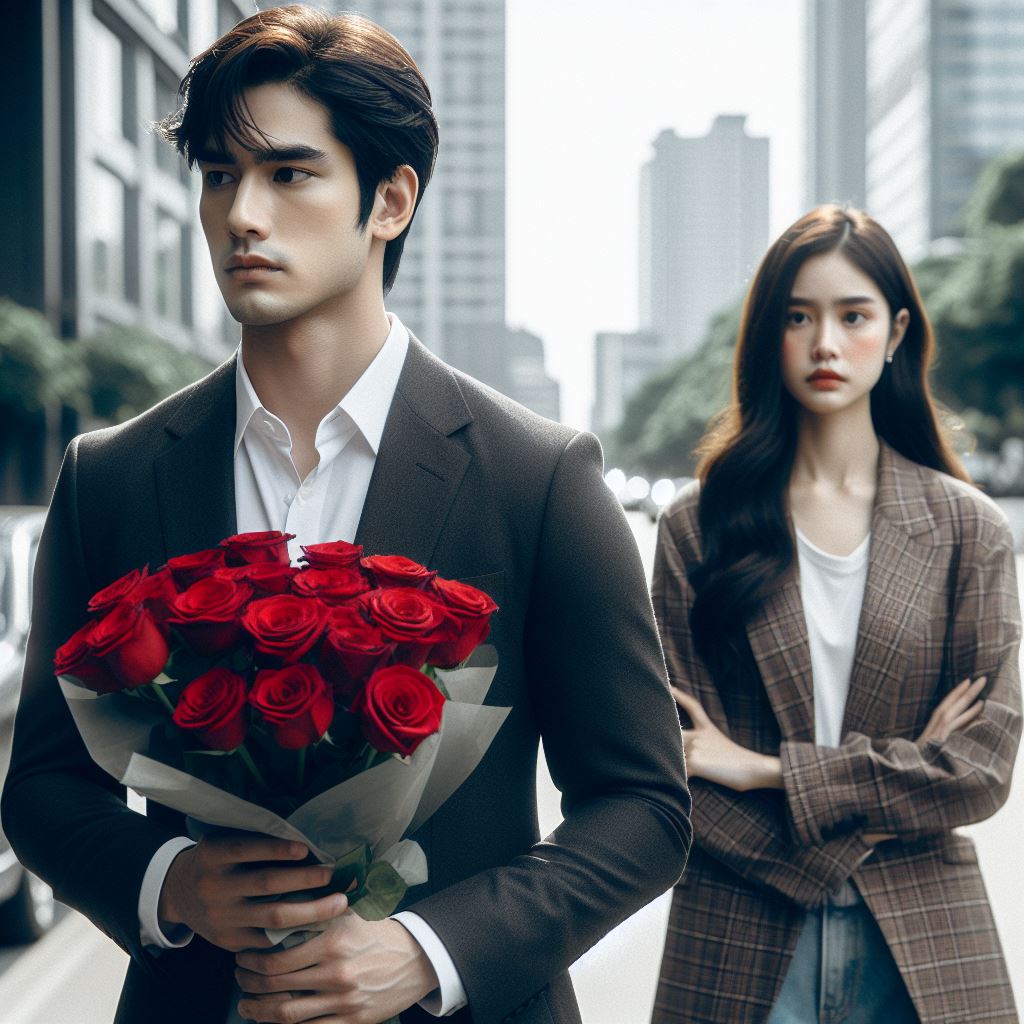}
\caption{man leaving woman}
\label{orientation2}
\end{subfigure} 
\caption{
Examples of images where the ground truth labels depend on spatial relationships between objects: \textit{distance} (a, e), \textit{inside/outside} (b, f), \textit{order} (c, g), \textit{orientation} (d, h). The images were created with the assistance of DALL-E 3.}
\label{fig:examples}
\end{figure*}

\subsection{Spatial Context}
\label{sec:spatial_context}
In this type of context, we would like to distinguish two categories: \textit{spatial relationships} and \textit{neighborhood}.

\subsubsection{Spatial relationships}
In this case, ground truth labels depend on the spatial relationships between objects i.e. \textit{distance}, \textit{inside/outside}, \textit{order}, \textit{orientation} (some illustrations are in Figure~\ref{fig:examples}). We~provide below illustrative real-world examples.

\paragraph{Distance} The fact that someone is standing within some distance from a shovel does not make this person a construction worker. The fact the man is holding a shovel increases the chances of him being a~construction worker. Therefore, the distance is a key. It applies to different occupations and their attributes. Consider that someone is wearing a stethoscope (most probably a physician) and the other case when someone is touched only by a small part of the stethoscope during an examination (patient) (see Figure~\ref{distance1}, \ref{distance2}). 
Another example is already said rollerblades -- not only there should be shoes and wheels in the picture but they should be properly attached.
Similarly, an image can be classified as a bump if the cars are in contact otherwise if there is a~significant distance between them, it is not a bump. This pairwise relationship can be extended to many objects i.e.~when there is an image of some people spread far away from each other, we cannot classify the scene as a crowd unless people are standing close to each other. 

The distance can have also a more physics-informed aspect. Consider there is an image of a squared table where all the legs are attached to the same side of a table. Such a table would collapse even though it has the same elements as the typical table (table top and four legs). 
Note that even if the model will correctly distinguish that such a sample is outside of a distribution of the real-world tables, we would not be able to correctly explain the model's decision using current XAI methods. 

\paragraph{Inside/ Outside} Suppose there are flowers in a vase. This is true only if the flowers are inside the vase, the close distance between the two objects is not enough (Figure~\ref{inside1},~\ref{inside2}).

\paragraph{Order} Suppose there is a train near a tunnel. If the order of the objects is (1) railway engine, (2) carriage, (3) tunnel, it means that a train is leaving the tunnel but if it is (1) carriage, (2) railway engine, (3) tunnel, then a train is entering the tunnel (Figure~\ref{order1}, \ref{order2}). Also, when we think of a hamburger, we expect a piece of meat between two pieces of bread. If there are two pieces of bread and a piece of meat on top, this is not a proper hamburger.

\paragraph{Orientation} There are two cases regarding orientation: an~angular position between objects (when we assign some coordinate system to each object) and the situation when objects are heading or standing back towards each other. The example for the latter is a man giving roses to a woman (they face each other) vs. a man holding roses but standing back to a woman (Figure~\ref{orientation1}, \ref{orientation2}). It may seem a subtle case of \textit{order} relationship because it can be translated to the question of whether the roses are between the two people or not, but we decide to treat it as a sub-case of orientation. The other more severe case is whether a person with a knife/gun is heading towards the other person or is going in the other direction. In the first case, there is a high chance that the image depicts `violence', whereas in the latter -- it is more probable that there is `no violence' in this particular frame. 

\subsubsection{Neighborhood}
The spatial context can be also understood in a manner more similar to the way it is treated in NLP where it is analyzed how many words surrounding the key part of the text we should input to the system. We would like to distinguish neighborhood as a separate subcategory of spatial context which can be thought of as \textit{co-presence} but within a strictly constrained location (spatial aspect). 

\subsection{Scale Context}
It is difficult to assess the size of a given object if there is no reference object of a known size in the image. This may impact the correctness of a classification. Suppose a photo of a silver hoop -- without contextual information, it can be either a ring or a bracelet. 
\\
In the following sections, we focus mainly on spatial context.

\section{Application Domains}
\label{sec:application}

Spatial context plays a significant role in many domains.
\paragraph{Street surveillance systems}
The videos from the streets and parks should be analyzed in an automated way to detect danger and alarm the police. 
The behavior of such AI systems should be trustworthy and safe (in compliance with regulations such as the EU AI Act and the Executive Order on the Safe, Secure, and Trustworthy Development and Use of Artificial Intelligence signed by President J. Biden). When analyzing situations on streets, the case with a human holding a knife/gun in a relative position to another human may occur. Therefore, the assessment of how severe the situations are, may depend on the orientation of the people towards each other.

\paragraph{Autonomous cars}
Safety is one of the biggest concerns in autonomous cars and so the same law regulations as in the case of surveillance systems apply. Suppose an autonomous car is approaching a crossing where a person is waiting to cross the street but there are no traffic signals. The vision system in an autonomous car identifies a human and a stick somewhere in the scene. Just based on this information, it is difficult to say whether the person is elderly and is holding a stick to facilitate walking or if it is elderly and a visible stick is just a part of a street sign or street lights. In the former, the car should predict that the needed time for the pedestrian to cross the street will be longer than in the latter and therefore, adjust its speed accordingly. Such adaptive cruise control can already benefit from the information about the surrounding cars to set the appropriate speed~\cite{adaptive_car} and so should be done also in the case of pedestrians. 


\paragraph{Healthcare} From the clinical point of view, it is important to check if lesions are clustered (primary) or widespread (with metastases) as it directly translates to the assessment of the stage of the disease and impacts the choice of treatment methods.
For instance, in histopathology, the data is saved in the form of Whole Slide Images (WSIs) which are of huge resolution showing the whole cuff-out lesion at the level of individual cells. Therefore, the processing of them using Deep Learning can be troublesome. 
There are two possible approaches: (1) analyze only separate patches taken out of the whole WSI (if local labels are available), which leads to a loss of broader contextual information, (2) analyze the~whole lesion in one go to preserve spatial context (tissue-level perspective)~\cite{clam,transmil,gigapixels_hist,protomil}.  

\paragraph{Land analysis (agriculture, environmental studies, archaeology)}
The satellite images are similar to the histopathological ones as they are often also of a very high resolution and so context may play an important role. The~satellite images can be used to assess the parameters of the soil, detect methane leakage and identify some historic ruins hidden in the soil. In these cases, the neighborhood context may be important during analysis with the help of DL models~\cite{satellite,land_cover}. 

\begin{figure}[!ht]
  \centering
\begin{subfigure}[b]{0.35\linewidth} %
\centering
\includegraphics[width=0.92\linewidth]{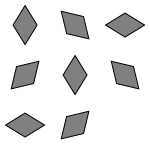}
\caption{input RPM}
\label{fig:test}
\end{subfigure} 
  \hfill
  \begin{subfigure}[b]{0.58\linewidth} %
    \centering
    \includegraphics[width=0.92\linewidth]{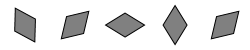}
    \caption{set of possible answers}
    \label{fig:ans}
    \end{subfigure} 
  \caption{Visual IQ test as an example of the task where spatial relationship understanding is required to solve it properly by choosing one answer from the set of given (b) to fill in the input RPM~(a).}
\end{figure}

\paragraph{IQ tests}
Spatial context is important also in the case of synthetic data such as visual IQ tests in the form of Raven Progressive Matrices (RPM)~\cite{rpm} where a test-taker is given a $3\times3$ grid of panels with one field missing (Figure~\ref{fig:test}) and the task is to choose the correct panel (matching the underlying relationship in the matrix) from the set of given panels (Figure~\ref{fig:ans}). The relationships within the images in the panels can occur within rows, columns and/or diagonals of the matrix. An example of the underlying rule within the RPM can be that there is a diamond shape in each panel that is rotated by the same angle from one panel to another one in each row but the starting orientations of the diamonds in the first column are different. Therefore, the RPM has to be analyzed as a whole, paying attention to spatial relationships between the objects in panels. For a~throughout survey on the application of Deep Learning to abstract reasoning tasks (including visual IQ tests) consult a~work of~\citet{rpm_survey}. 

\section{Spatial Context in DL Models}
Spatial context can be valuable when performing Deep Learning tasks. We analyze whether the information on spatial aspects is concerned within the design (architectures) and training schemes of DL models. 

\paragraph{Model architecture families}
In the early days of neural networks, the images were flattened before being passed to multilayer perceptrons, which led to an irrevocable loss of spatial aspect within the data. In order to overcome this downside, convolutional neural networks (CNNs) were proposed that analyze the images as matrices. Such a design of convolutional layers was supposed to capture local patterns. 
Later, a new architecture, Vision Transformer, was introduced with a self-attention mechanism as a main part. The self-attention is supposed to retrieve even the long-range relationships between the input patches. Additionally, the~model is fed with patch positional encodings.

\begin{figure}[h!]
\centering
\includegraphics[,width=0.66\columnwidth]{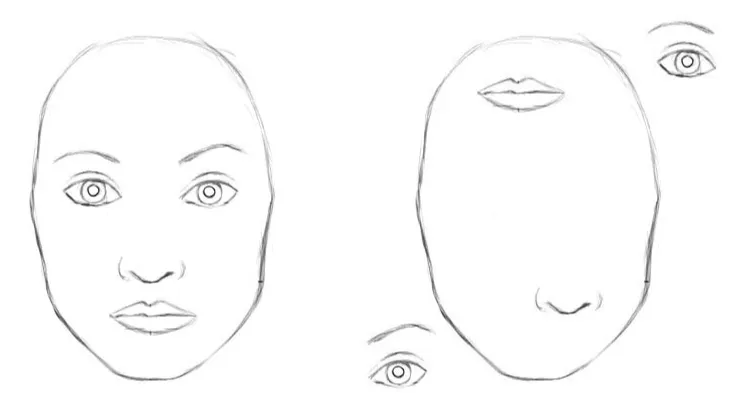}
\caption{In the two images, there are the same elements so a CNN most probably will classify them as a face in both cases unlike a CapsuleNet that will not classify a \textit{deformed face} as a \textit{face} as spatial relationships between elements are not properly preserved
\cite{face}.}   
\label{fig:face}
\end{figure}

Within this timeline of consistent progress in the field, two solutions outside of the main track deserve attention: CapsuleNet~\cite{capsulenet}, MLP-Mixer~\cite{mlpmixer}. 
The motivation behind the idea of CapsuleNets was to propose an architecture that will take into account the spatial relationships between objects in the image as opposed to CNNs that were claimed to pay attention only to different elements of the image without considering spatial context due to continuous downsampling operations~\cite{capsulenet_survey}. Such a by-design behavior of CNNs was perceived as a drawback. The illustrative example is provided in Figure~\ref{fig:face}.

Another worth-mentioning solution that went off the beaten track was MLP-Mixer. The authors regardless of the trend to benefit from contextual information, showed in one of the experiments, that MLP-Mixer is invariant to the order of input patches and pixels within the patches treating it as an asset. The performance of the MLP-Mixer was the same on the said modified images as on the original ones. In the case of ResNet, the performance on the images where the spatial relationships were corrupted in the said manner, dropped significantly. It is claimed that it is due to CNNs' strong inductive bias.

\paragraph{Context-oriented solutions}
\label{sec:model_context}
Some solutions were designed to benefit directly from contextual information within images. The context is expected to help in emphasizing the required features while suppressing the undesired variation which is especially important when there are various orientations and scales of the same objects~\cite{land_cover}.

The topic was analyzed for natural scene images (mostly in a detection task). The early approaches are summarized in a~survey~\cite{context_survey}. The vast number of works incorporated the concept of neighborhood into the pipeline. One of those is spatial-context-aware deep neural network~\cite{spatial_multiclass} used for multi-label classification when bounding boxes are provided. In the solution, there are two bounding box generation branches -- the object and context ones. The first one localizes precisely the objects (tight bounding boxes) whereas the second expands the bounding boxes by incorporating the neighborhood around the object. Later the features are extracted from the two branches and combined to perform a final classification benefitting from contextual information. 

Another approach to address the topic of spatial context in models is to use Feature Pyramid Networks~\cite{pyramid} which process the images at different scales to gain a~better understanding of global and local context. \\Domain-specific solutions:
\begin{itemize}
    \item Histopathological data: The concept of incorporating information from various resolutions was a central part of several works~\cite{gigapixels_hist, spatial_random_fields,multiscale_branch}. Another approach was to use information from neighboring patches of the one that was under analysis at a particular moment. ~\citet{near_patches} propose two 2D-LSTM spatial context branches on top of a CNN feature extractor to obtain continuous and discrete spatial context-dependent information. 
    \item Remote sensing data:
    Two main challenges are identified when trying to benefit from global and local context in land cover analysis: the ambiguity of global context and lack of efficient context combination strategy~\cite{land_cover}. To mitigate these limitations,~\citet{land_cover} propose a mechanism of a fusion of global and local contexts under the guidance of uncertainty. The access to the information on neighborhood pixels is provided to the system in work~\cite{satellite}.

\end{itemize}

\paragraph{Pretraining}
The spatial aspect was also incorporated into model pretraining schemes. 
One class of such approaches is inpainting where the model is asked to fill up the missing parts of the input image. In such cases, the generation of missing image parts is conditioned on its surroundings~\citep[i.e.][]{inpainting, mae}. The resulting context encoder can be later used for various downstream tasks as they capture the semantics of visual structures. 

The pretraining scheme where the idea of spatial context was incorporated in a more explicit form is described by~\citet{pretraining_spatial_grid}. The model is given two neighboring patches from an image and is asked to output their spatial configuration out of 8 possibilities. There is a $3\times3$ grid where the first patch is set and the position of the second within the~grid is a target.

\paragraph{Context relationships as output}
In all of the aforementioned works except the one by~\citet{pretraining_spatial_grid}, the~understanding of spatial context by DL models is not the goal itself but a means towards better performance in the main task. In~the~work of~\citet{fashion_mnist}, the prediction of spatial context is the center of the experiments. In~the~first experiment, the model is given an image of two cloth pieces and is supposed to classify each object and output the type of spatial relationship that is present (\textit{left, above, below left, and below right}). In the second experiment, one sample input is an image containing three objects and one-hot embeddings of two objects. The task is to output the spatial relation between the two objects mentioned within the input embeddings. \\
The spatial relationships within the~output are also investigated in generative models when relative positions of the objects are specified in the prompts~\cite{diffusion_spatial}.

For another view on the topic of context understanding in computer vision consult a survey~\cite{dl_context}.

We acknowledge that there are some recent solutions dedicated to scene understanding i.e. VisProg~\cite{visprog}, ViperGPT~\cite{vipergpt} and physics-informed NNs~\cite{pinns} that potentially could understand that the right position of the legs in the table prevents collapse.
This shows that the topic of spatial context is recognized as important in the community working on DL models. However, we show that it is rather overlooked in the XAI community. These recent DL solutions are not strictly vision models, which are of the biggest interest in the XAI community where the focus is on investigation of the mature solutions like CNNs and ViTs that already find real-life applications.

\section{XAI Failures}
\label{sec:failures}
We provide examples when popular XAI techniques fail to explain in a comprehensive way the predictions of the model when spatial context is crucial. We fine-tuned \textit{Resnet-50} and \textit{ViT base} models in low data regime as it is done in work~\cite{vpt} using datasets from Visual Task Adaptation Benchmark (VTAB)~\cite{vtab}. The~models were pretrained on Imagenet: \textit{Resnet-50} -- in a supervised manner, \textit{ViT base} -- in a contrastive manner (\textit{Moco v3}). We focused on the subset of VTAB called `structured' where the labels depend on spatial context. In the experiments, we use two datasets: \textit{KITTI}~\cite{kitti} where images were collected using sensors in the car -- the task is to predict the binned distance to the closest vehicle in the scene, \textit{dsprites}~\cite{dsprites} where images of simple shapes undergo rotations and other shifts in the space -- the task is to predict binned orientation. Hence, we analyze the only real-life dataset and one of the few synthetic datasets in VTAB.

\begin{figure*}[hbt]
\centering
\begin{subfigure}[b]{0.48\linewidth} %
\centering
\includegraphics[width=0.93\linewidth]{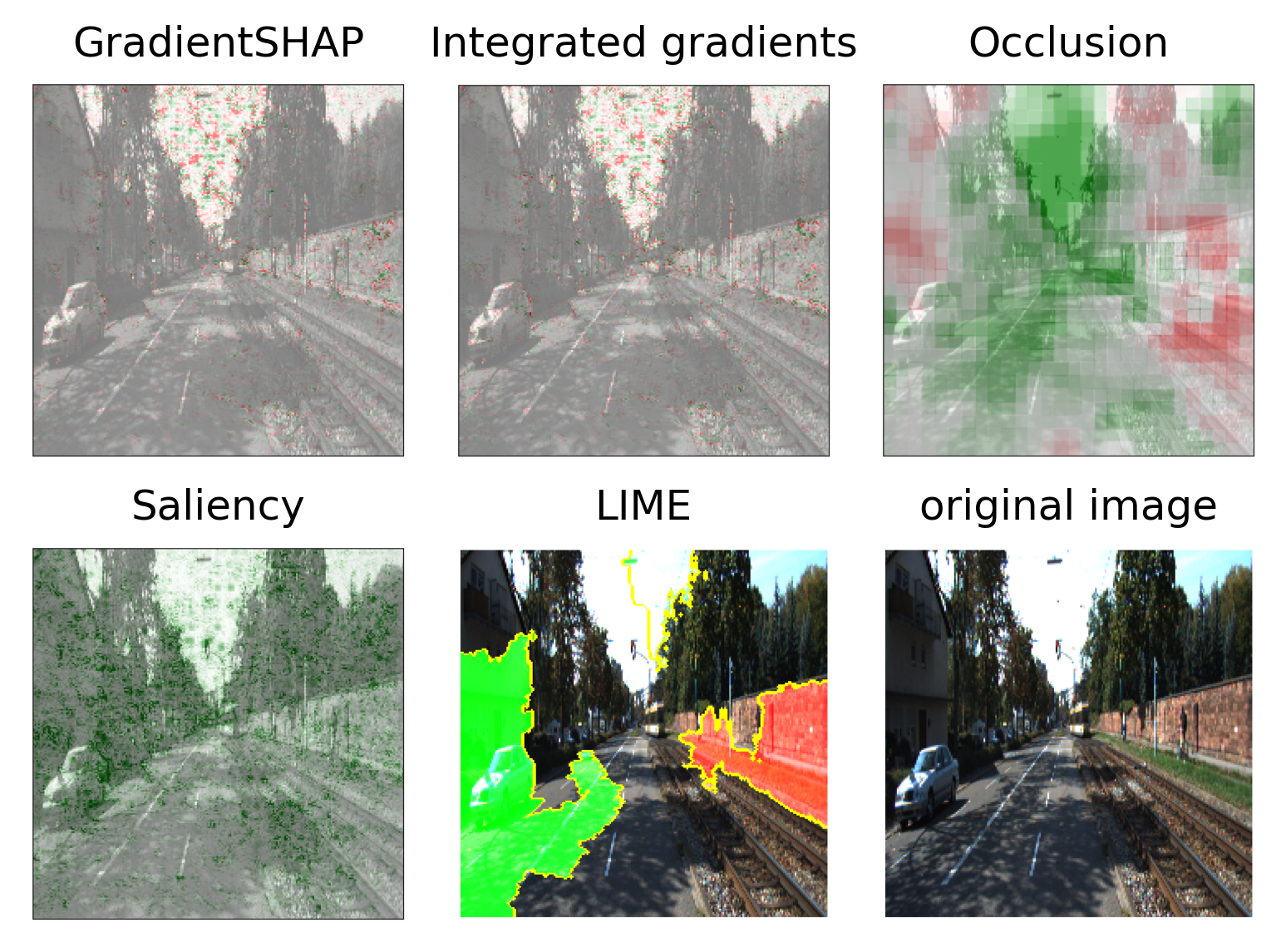}
\caption{\textit{KITTI} dataset, \textit{Resnet-50}}
\end{subfigure} 
\begin{subfigure}[b]{0.48\linewidth} %
\centering
\includegraphics[width=0.93\linewidth]{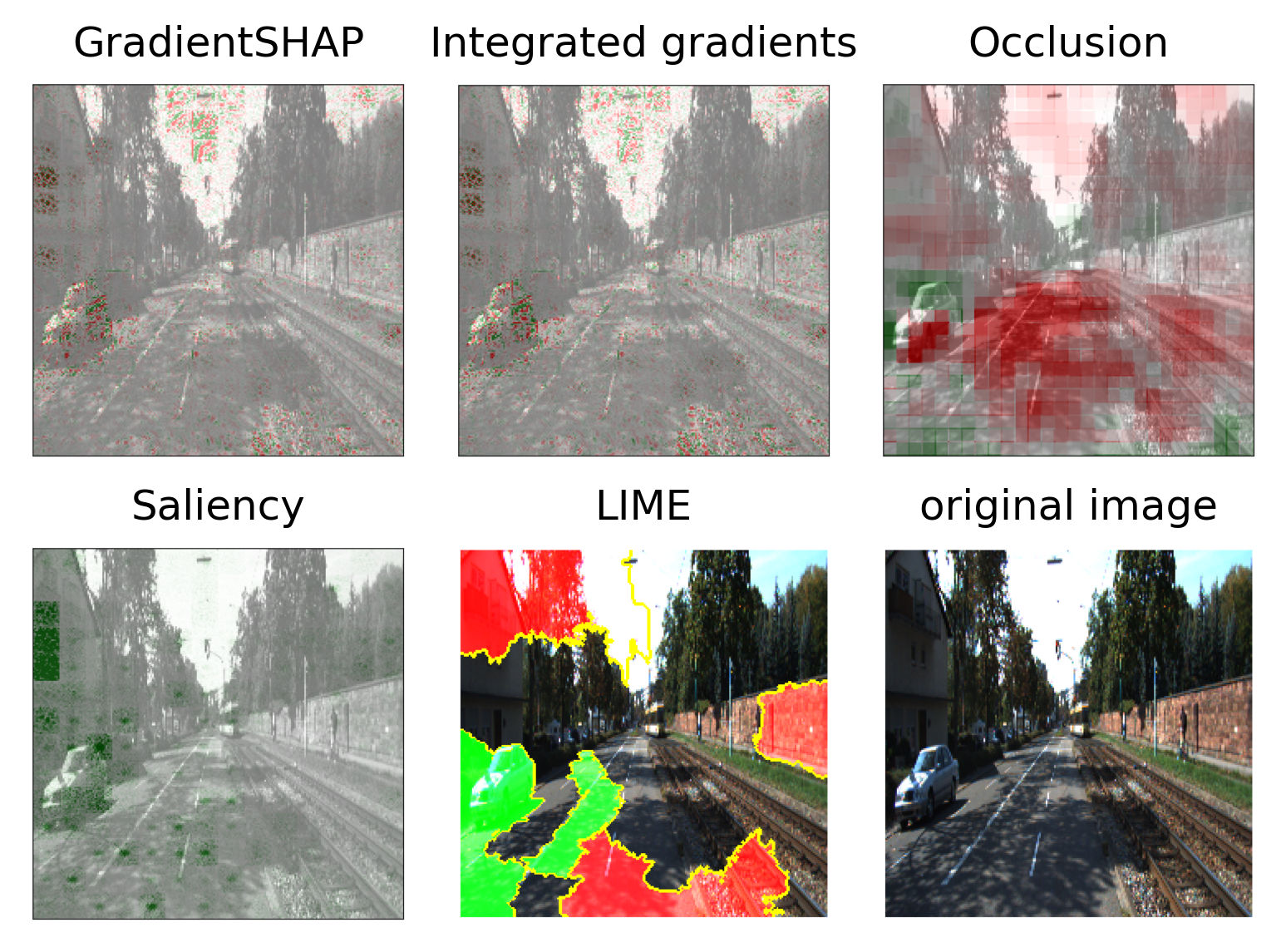}
    \caption{\textit{KITTI} dataset, \textit{Moco v3}}
    \end{subfigure} 
\begin{subfigure}[b]{0.48\linewidth} %
\centering
\includegraphics[width=0.93\linewidth]{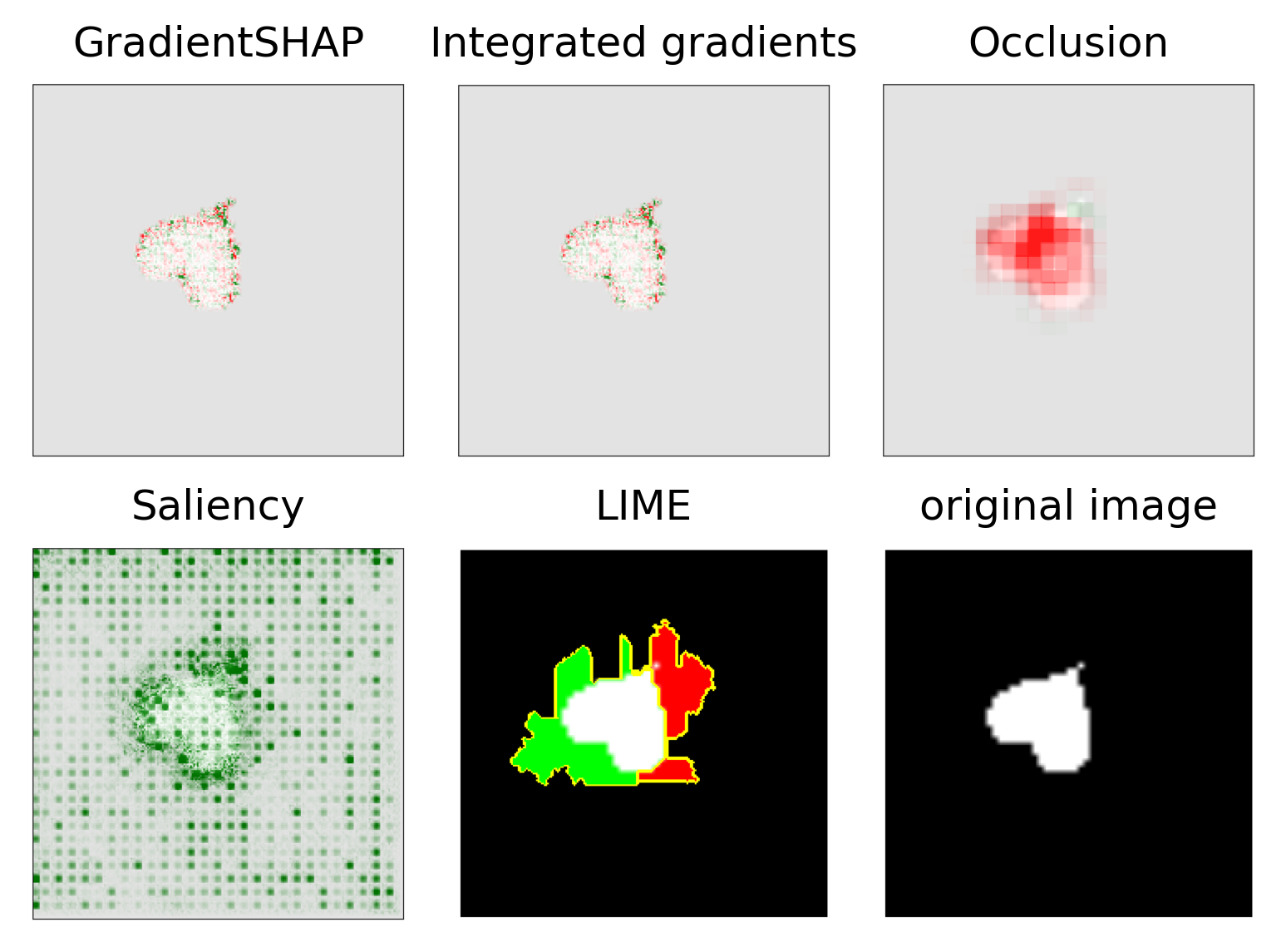}
\caption{\textit{dsprites} dataset, \textit{Resnet-50}}
\end{subfigure} 
\begin{subfigure}[b]{0.48\linewidth} %
\centering
\includegraphics[width=0.93\linewidth]{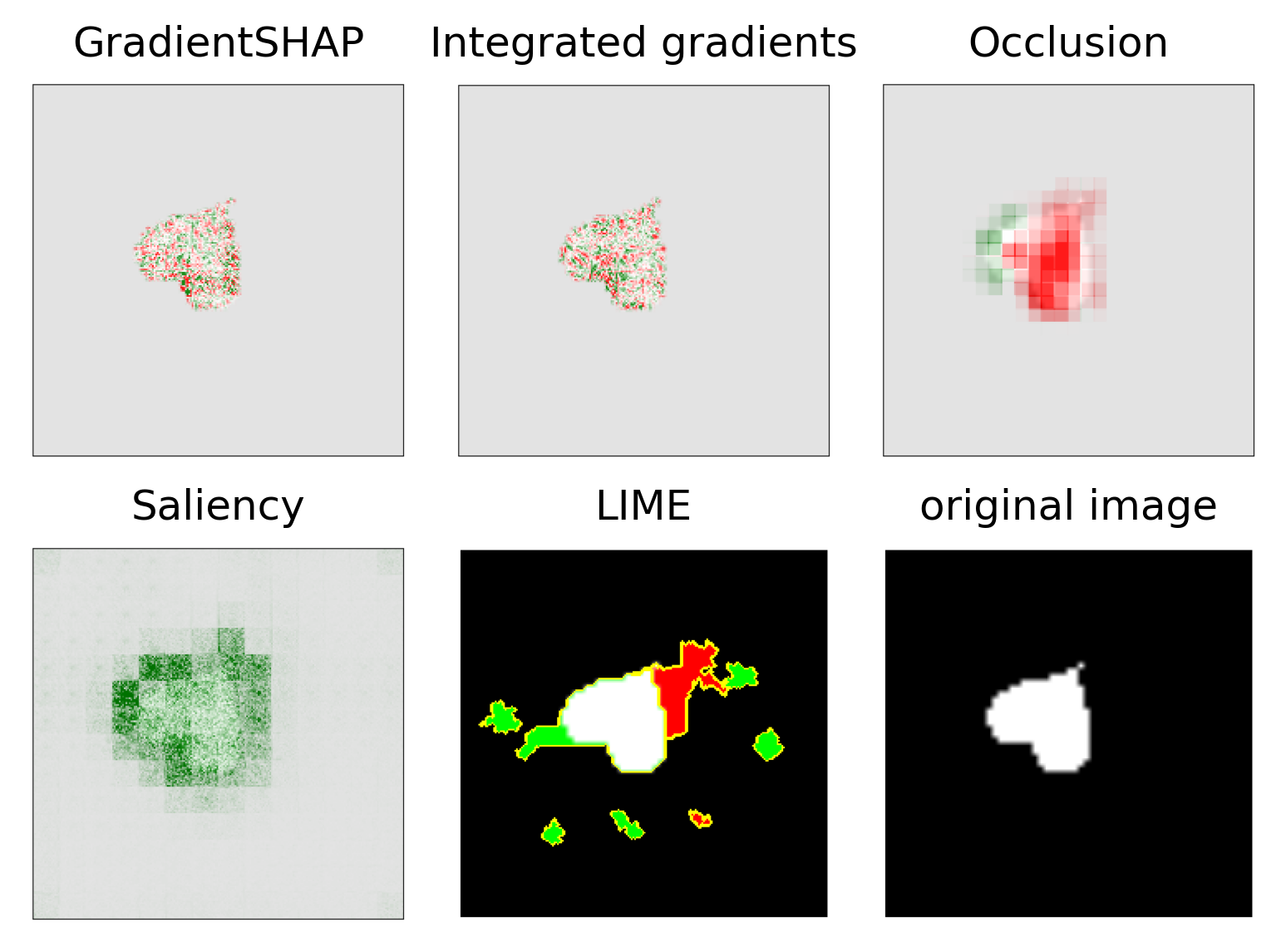}
    \caption{\textit{dspites} dataset, \textit{Moco v3}}
    \end{subfigure} 
\caption{The explanations of correct predictions of \textit{Resnet--50} and \textit{Moco v3} on the samples from the \textit{KITTI} and \textit{dsprites} datasets. In~the~first case, the task is to predict a binned distance to the nearest vehicle, whereas, in the second, to predict a binned orientation of a~shape. Five popular XAI methods were used for the study. The color spectrum from red to green depicts the extent to which a~particular image part contributed to the model’s
prediction (from negative to positive impact).}
\label{fig:failures}
\end{figure*}

Having fine-tuned models of a satisfactory performance (similar to the one claimed by~\citet{vpt}), we applied 5 popular XAI techniques: GradientSHAP~\cite{grad-shap}, Integrated gradients~\cite{ig}, Occlusion~\cite{occlusion}, Saliency~\cite{saliency}, LIME~\cite{lime} to check if they manage to explain the model’s decisions correctly. In Figure~\ref{fig:failures}, we provide images with the models' explanations. The colors reflect the extent to which particular parts contribute to the final model's decision (green -- means positive attribution, whereas, red -- negative). We can see that the existing popular XAI methods fail to explain the correct model decisions i.e. that the nearest vehicle is within the~distance of 8 to 20 meters.

The explanations fail -- they are inaccurate, vague and difficult to interpret (Figure~\ref{fig:failures}). It can be seen that the nearest vehicle is highlighted in green (using the LIME method), however, it is not clear that the distance is crucial. One may conclude that the same explanations could also be seen in the object recognition tasks i.e. cars vs. planes. Therefore, it may be important to design spatial XAI techniques that would be tailored to different ML tasks or context types.

In the XAI field, the visual explanations are the most popular. However, it seems that they may not be sufficient to properly highlight the role of spatial relationships in predictions. Hence, some other approaches are needed.

\section{Spatial Context in XAI}
\label{sec:xai}
The topic of spatial context was considered when training neural networks but it was of much smaller interest in XAI. We showed that there is still plenty to do as popular XAI methods fail. Note that without proper XAI tools, we would not be able to explain the predictions of CNNs and CapsuleNets when the same features are within input images but placed differently in the scene. The existing methods in the case of \textit{deformed face} (Figure~\ref{fig:face}) will most probably highlight the same key elements of the image both in the~case of CNNs and CapsuleNets even though the two DL models will return different predictions. 

The investigation of whether the spatial context plays a role and to what extent in the decision-making of DL models is a niche that should be fulfilled. 
We call this area of research \textit{spatial XAI}. There are only a few works in this domain. As~a~result of a survey, we distinguish four main approaches: \textit{intrinsically explainable models}, \textit{methods introducing measures of spatial context}, \textit{leveraging current XAI methods} and the ones \textit{based on input-output relationships}.

\paragraph{Intrinsically explainable models}
The work~\cite{context_detection} encompasses two aspects as it proposes a new explainable-by-design architecture for detection and the metrics to evaluate the importance of context.
The authors modify the well-known detector architecture by adding a contextual learner to extract contextual cues from the scenes. The contextual feature maps for each class give insights into the way the model benefits from context.
The other contribution is the two scoring functions to measure the contextual relevance of detections in relation to other objects present in the image
and the general scene. The first score is calculated as a summation of the elements of the feature map corresponding to a bounding box within the mask region, divided by the area of the bounding box’s mask. The second scoring method is analogous, but it is calculated based on contextual feature maps generated for each class instead of bounding boxes. The metrics are later used to compare different DL models. The authors challenged the proposed DL model on a set of hypotheses referring to spatial context. It~was possible thanks to the fact that they had control over the generation of a synthetic dataset so that you could prepare samples violating the underlying spatial rules in the dataset and see how the model behaves in such scenarios. 

\paragraph{Measures of spatial context}
\citet{proto_spatial} analyze the spatial aspect within explainable-by-design prototypical models. They define the problem of \textit{`spatial explanation misalignment'} and introduce a metric to quantify this phenomenon. Moreover, they propose a benchmark and a method to mitigate this spatial misalignment in this particular kind of models. It was observed that sometimes the location of the explanation within the input image changes when the input image is modified in non-meaningful regions that should not influence the model prediction (distractors). 

\citet{dscon} focuses on metrics to evaluate the preservation of spatial context in attention-based Multiple Instance Learning vision models that most often find application in the analysis of high-resolution images. The authors proposed three spatial context measures ($SCM_{features}$, $SCM_{targets}$, $SCM_{residuals}$). The quantitative measurement is possible thanks to the incorporation of spatial regression models which are run on patch features to predict the respective attention scores assigned by the DL models. In the work, the histopathology use case is provided where the spatial context in the input is understood as spatial autocorrelation between patches of Whole Slide Images. 

\paragraph{Leveraging current XAI methods} 
One of the early XAI methods is LIME~\cite{lime} which operates on superpixels from input images assigning them the importance score explaining the model's decision. In work, \cite{lime_location}, the authors propose to build upon the LIME method by incorporating logic rules obtained by the Inductive Logic Programming system, Aleph. The authors, therefore, propose to combine visual and symbolic methods to explain spatial relationships.

\paragraph{Input-output relationships}
In Deep Learning, often ablation studies are performed where the particular parts of the networks are switched off to check whether they are necessary and what their role is. Such an evaluation is often performed to compare the performance of the original network and the modified one. In XAI, a similar idea is behind permutation-based methods where parts of the input image are erased to see how it impacts predictions (i.e. LIME). In work, \citet{pcam}, an analogous approach was used to evaluate the importance of neighborhood. The~model used to predict whether a tissue patch contains a tumor or not was trained on images containing the region of interest and neighborhood. The~ground truth label is provided only based on the region of interest. Later, inference was performed using images with a gradually decreased size of the neighborhood. It was analyzed how much the initial performance metrics were impacted by the limitation of contextual information. In the case of a so-defined neighborhood, a simple application of the existing XAI techniques could show whether models pay attention to the neighborhood or not.

Another input-output relationship approach accompanied by a new XAI method is proposed by~\citealp[]{ablation_spatial}.
The authors cover some parts of the input image (using a~sliding window) and later compute the influence it has on a value of cross entropy compared to the situation when no modifications to the input image are made. As a result, the heatmaps showing key parts of the images are generated. The authors present some heatmaps visualizing the relationships (\textit{below}, \textit{behind} and \textit{above}). However, the limitation of the provided heatmaps is the fact that without the authors' captions, it would be difficult to decipher what type of relationships are highlighted. In the work, the authors also zero out different groups of neurons in MLP layers to check if they are responsible for detecting particular spatial relationships. 

The analysis of spatial relation understanding is also performed in text-image generation models where it is assessed if the spatial relations specified in the prompt are preserved within generated images~\cite{paintskills}. 

\section{Needed Research Directions}
\label{sec:directions}
In some of the mentioned papers on spatial context~\cite{ablation_spatial,context_detection, fashion_mnist, lime_location}, the claims were investigated on synthetically generated datasets in order to have better control over the experiment setup which facilitates the verification of the research hypotheses. 
These datasets were not made publicly available and, moreover, were not of sophisticated compositionality as there were only some simple objects in relative relationships put on the plain background. With the advent of diffusion models i.e. stable diffusion~\cite{stable_diffusion}, it is possible to generate more realistic images with the desired spatial relationships between objects (samples in Figure~\ref{fig:examples}). Therefore, we postulate for new datasets reflecting scenarios where ground truth depends on spatial relationships. 

\paragraph{Assessment of spatial reasoning skills in text-to-image models} PAINTSKILLS~\cite{paintskills} is a compositional diagnostic evaluation dataset where images are generated using prompts containing information on spatial relationships (\textit{above, below, left, right}) between objects. Note that these relationships are relatively simple compared to the ones we outlined in Section~\ref{sec:spatial_context}. The dataset is used as one of the scenarios verifying reasoning skills within the recently proposed benchmark, Holistic Evaluation of Text-to-Image Models (HEIM)~\cite{holistic}. Within the reasoning scenario, it is checked whether models understand objects, counts, and spatial relations (compositionality) within images. Some task-specific benchmarks dedicated to the evaluation of the vision-language model's ability to understand scene composition were proposed~\cite{crepe,cola}. It seems that the understanding of compositionality within text-to-image solutions is assessed only at a high level, however, in XAI the focus is on a deeper analysis of a model decision-making process. Here, we could conclude that the recent approaches are more about checking the accuracy of the model in outputting what is expected instead of tracking the model's reasoning process.

\paragraph{Need for spatial context benchmark for non-generative models} We believe that analogous or even more in-depth benchmarks should be proposed to verify the reasoning of purely vision models with a focus on assessing whether the spatial context is properly taken into account. To the best of our knowledge, there is no such benchmark even though there are many vision models that benefit from contextual information (Section~\ref{sec:model_context}). 
 The analysis of some reasoning capabilities is performed in the case of models solving IQ tests but not in the real-world cases that we point out.  
 
 \paragraph{Need for measures} A benchmark incorporating spatial relations understanding could serve as an important point in a checklist before the launch of DL models in critical domains. This could fulfill some of the requirements of ensuring the safety of models.
 For this benchmark to be successful, some new metrics to evaluate to what extent the models take into account spatial context should be proposed.
 So far models' investigation mostly relies on visual XAI methods (heatmaps) which are not a sufficient way to explain models in cases when spatial context matters. 
 
\paragraph{Need for diverse XAI methods} One of the interesting directions seems to be incorporating new types of explanations on top of visual ones (as it was done by~\citet{lime_location} with symbolic rules) or adding them as a separate element. Moreover, it may be valuable to focus on studying the inner workings of DL models and analyzing them using some probe tasks~\cite{probe}. The motivation behind probe tasks is that if we can train a probe model on top of some internal representation (extracted from the DL model) to correctly predict a property of the input data, it means that the property is encoded somewhere in the representation. 
Overall, we should not only focus on qualitative analyses but also try to enrich them with quantitative ones for a more holistic understanding of vision models' decisions. In Appendix~\ref{sec:appendix}, we discuss ideas on how to incorporate spatial XAI into a Data Science pipeline.

\section{Conclusions}
\label{sec:conclusions}
Recently, in the XAI field the shift of the paradigm from \textit{where} to \textit{what} was proposed, meaning that instead of only highlighting some parts of input images, we should also be able to name them semantically. 
We propose to go further and focus on the notion of \textit{how} the objects within the images are oriented towards each other (contextual information).
We provide many examples where the ground truth labels depend on the spatial relationship between objects i.e. in autonomous cars, surveillance systems, healthcare, agriculture, and environment studies. In these cases, the existing XAI methods will fail to provide valid explanations of the predictions of the DL models. It may happen that despite having two different models' predictions for two images, we would get semantically the same explanations. We provide examples of failures of XAI methods (unclear explanations).

We conducted a survey on papers that focused on the topic of spatial context within DL models and XAI techniques. It~turned out that there are only a few works in so-called \textit{spatial~XAI} compared to the number of DL models benefitting from spatial context. In this work, we would like to direct the attention of the XAI community to this largely unexplored area. 
 We propose some potential research directions in the field of spatial XAI like a holistic benchmark that will evaluate models from the spatial context point of view and allow for quantitative assessment. We believe that by inspiring a community to address the topic of spatial XAI, it is possible to move forward research in XAI in general, as it would help in going out of the realm of current development directions. Moreover, a better understanding of how the models \textit{'see'} spatial context may lead to better DL models' design in the~future.

\section*{Acknowledgements}
We would like to thank the reviewers for the useful feedback that enriched the final version of the paper. 
This research was funded by the Warsaw University of Technology within the Excellence Initiative: Research University (IDUB) programme. The work was carried out with the support of the Laboratory of Bioinformatics and Computational Genomics and the High Performance Computing Center of the Faculty of Mathematics and Information Science, Warsaw University of Technology.

\section*{Impact Statement}
This paper presents work whose goal is to advance the field of Machine Learning. There are many potential societal consequences of our work, none which we feel must be specifically highlighted here.

\bibliography{example_paper}
\bibliographystyle{icml2024}

\newpage
\appendix
\onecolumn
\section{Spatial XAI within Pipeline}
\label{sec:appendix}
The importance of spatial context within input data to DL models may not be black or white and some procedures may be needed when using spatial XAI in the future. There are many use cases where the spatial context is crucial. But there may be some exceptions. For instance, there can be an image of a room without any thermometer in it and we ask a model what the temperature in the room is. In this case, the context may not help in solving this task as simply visual information is not enough to solve this particular task.

We distinguish two possible scenarios of how to use the spatial XAI in the real-life Data Science pipeline but it may not be a~finite list: (1) check in advance if the spatial XAI is applicable, (2) critically analyze the output of spatial XAI techniques.

In the first case, we assume that the process will be that the evaluators of the DL systems will know whether, in the particular task/ dataset, the spatial context is crucial. In some cases (i.e. ML for healthcare), the~experts’ domain knowledge may be needed. 
This way it may be assessed if the use of spatial XAI techniques for the investigation of the model is justified. Such \textit{`check before use'} strategy is analogous to the case with the spatial regression models that can be applied instead of the~Ordinary Least Square regression only if the observations and the residuals are not independent and form some clusters. This is formally checked from the perspective of spatial autocorrelation using the Global Moran's I test. Maybe, a similar form of verification could be introduced to check if the use of spatial XAI method is justified in a particular case or not. 

However, in some use cases, it may not be obvious if the spatial context is meaningful. Hence, the scenario could be that we apply the spatial XAI techniques to every use case and then evaluate the output -- whether the spatial XAI methods show that indeed the spatial context plays a key role in the particular task. Such \textit{`post-hoc'} analysis can potentially bring some valuable insights into different phenomena where domain experts do not know a priori if the spatial context is important.


\end{document}